\documentclass[11pt,a4paper]{article}
\usepackage[hyperref]{acl2020}
\usepackage{times}
\usepackage{pifont}
\usepackage{latexsym}

\usepackage{balance}
\usepackage{helvet}  
\usepackage{courier}  
\usepackage{url}  
\usepackage{graphicx}  
\usepackage{amssymb}
\usepackage{amsmath}
\usepackage{algorithm}
\usepackage{algorithmic}
\usepackage{amsthm}
\usepackage{multirow}
\usepackage{pgffor}
\usepackage{graphicx}
\usepackage{epstopdf}
\usepackage{tikz}
\usepackage{epstopdf}
\usepackage{mathtools}
\usepackage{enumitem}
\usepackage{subfigure}
\usepackage{tabulary}
\usepackage{color}
\usepackage{url}
\usepackage{flushend}
\usepackage{stfloats}
\usepackage{tcolorbox}
\usepackage[utf8]{inputenc}
\usepackage[english]{babel}
\usepackage{tabularx}
\usepackage{CJKutf8}
\usepackage{microtype}
\usepackage{textcomp}
\usepackage{booktabs}
\usepackage{colortbl}
\usepackage{soul}
\usepackage{natbib}
\usepackage{appendix}
\aclfinalcopy

\definecolor{camel}{rgb}{0.72, 0.55, 0.14}
\title{Commonsense Evidence Generation and Injection in Reading Comprehension}

\author{Ye Liu$^1$, Tao Yang$^2$, Zeyu You$^2$, Wei Fan$^2$ and Philip S. Yu$^1$\\
$^1$Department of Computer Science, University of Illinois at Chicago, IL, USA\\
$^2$Tencent Hippocrates Research Lab, Palo Alto, CA, USA\\
\{yliu279, psyu\}@uic.edu, \{tytaoyang, davidwfan\}@tencent.com, youz@onid.orst.edu
}

\date{}

\begin{document}
\maketitle
\begin{abstract}
Human tackle reading comprehension not only based on the given context itself but often rely on the commonsense beyond. To empower the machine with commonsense reasoning, in this paper, we propose a \textbf{C}ommonsense \textbf{E}vidence \textbf{G}eneration and \textbf{I}njection framework in reading comprehension, named \textbf{CEGI}. The framework injects two kinds of auxiliary commonsense evidence into comprehensive reading to equip the machine with the ability of rational thinking. Specifically, we build two evidence generators: the first generator aims to generate textual evidence via a language model; the other generator aims to extract factual evidence (automatically aligned text-triples) from a commonsense knowledge graph after graph completion. 
Those evidences incorporate contextual commonsense and serve as the additional inputs to the model. 
Thereafter, we propose a deep contextual encoder to extract semantic relationships among the paragraph, question, option, and evidence. Finally, we employ a capsule network to extract different linguistic units (word and phrase) from the relations, and dynamically predict the optimal option based on the extracted units. Experiments on the CosmosQA dataset demonstrate that the proposed CEGI model outperforms the current state-of-the-art approaches and achieves the accuracy (83.6\%) on the leaderboard.



\end{abstract}

\section{Introduction}
Contextual commonsense reasoning has long been considered as the core of understanding narratives \cite{hobbs1993interpretation, andersen1973abductive} in reading comprehension \cite{charniak1990probabilistic}. Despite the broad recognition of its importance, the research of reasoning in narrative text is limited due to the difficulty of understanding the causes and effects within the context. Comprehend reasoning requires not only understanding the explicit meaning of each sentence but also making inferences based on implicit connections between sentences.
\begin{figure}[t]
\centering
\includegraphics[width=0.95\linewidth]{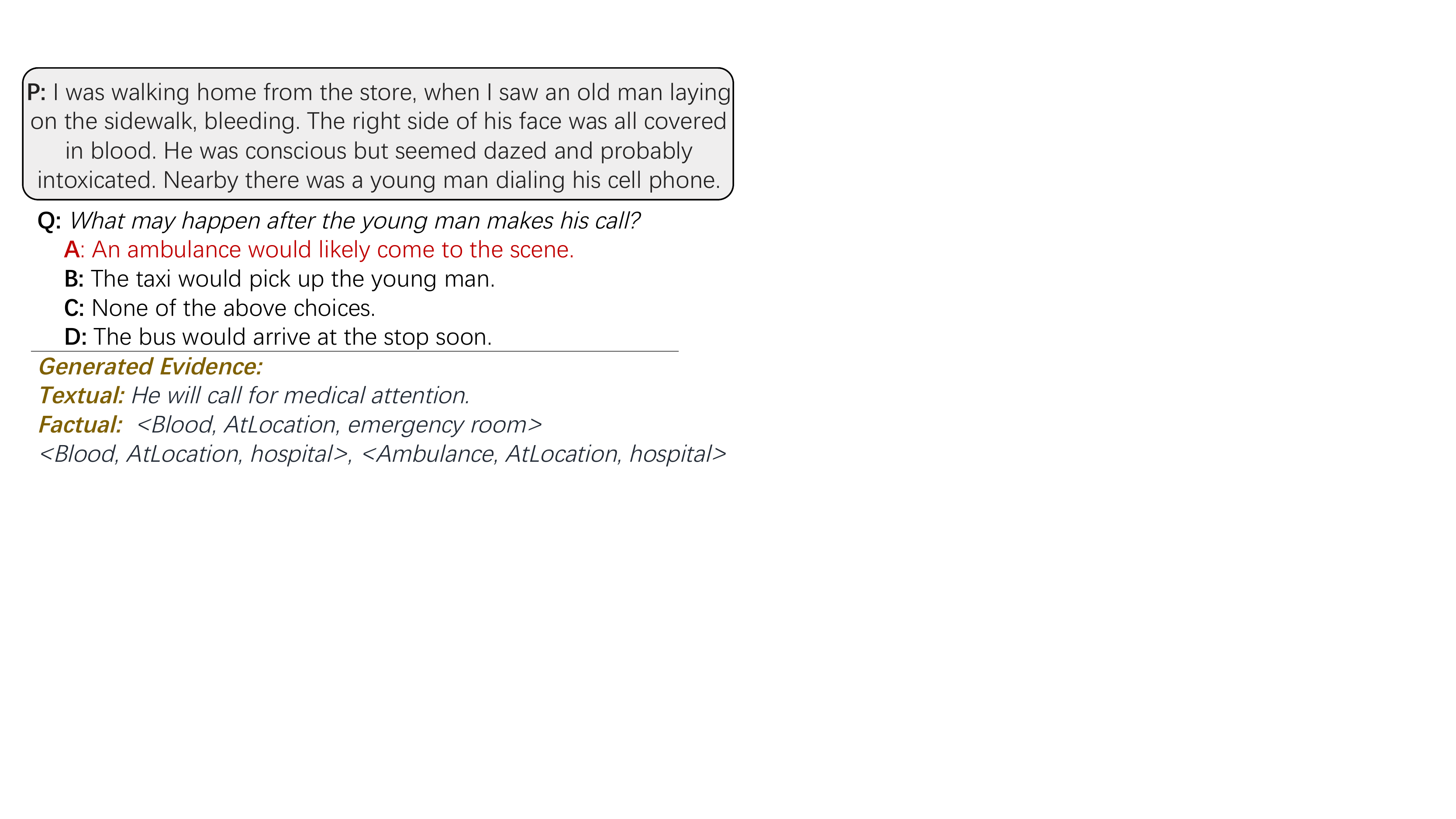}
\vspace{-0.7em}
\caption{Example of generated evidence can help answer the commonsense question.} 
\label{fig:Intro1}
\vspace{-1.3em}
\end{figure}

To answer a contextual commonsense question correctly, two important characteristics need to be well considered. First, the information that is required to infer a correct answer may be beyond the context, hence adding external commonsense knowledge to guide the reasoning is necessary. Second, how to use the external knowledge to gain contextual understanding between the paragraph, question and option set is difficult but important. Despite the great success of large pre-trained models such as BERT \cite{devlin2018bert}, GPT \cite{radford2018improving} and RoBERTa \cite{liu2019roberta}, recent studies suggest that those models fail to capture sufficient knowledge and provide commonsense inference. For example, \citet{poerner2019bert}, show the language models do well in reasoning about entity names, but fail to capture rich factual knowledge. Moreover, \citet{talmor2019olmpics} state that language models fail on half of the eight reasoning tasks that require symbolic operations such as comparison, conjunction \& composition. 

To this end, we introduce a \textbf{C}ommonsense \textbf{E}vidence \textbf{G}eneration and \textbf{I}njection framework in reading comprehension, named \textbf{CEGI}, which generates useful evidence from textual and factual knowledge and injects the generated evidence into pre-trained models such as RoBERTa. We propose to generate evidence regarding the facts and their relations. More specifically, we use language models to generate textual evidence and extract factual evidence from a knowledge graph after graph completion. We then inject both evidences into the proposed contextual commonsense reasoning model to predict the optimal answer. As shown in Figure \ref{fig:Intro1}, the \textcolor{camel}{\textit{Textual Generated Evidence}} ``He will call for medical attention'' and \textcolor{camel}{\textit{Factual Generated Evidence}} ``both blood \& ambulance locate at hospital'' can help the model find the correct answer ``An ambulance would likely come to the scene''. 

To capture the relations between the paragraph and question, many reading comprehension models \cite{zhang2019dcmn, tang2019multi} have been proposed. However, those models essentially reasoning based on the given context but without an understanding of the facts behind. 
In addition, in many situations, the candidate option set contains distractors that are quite similar to the correct answer. Therefore, understanding the relations among the option set is also important but not well-addressed in the current state-of-the-art models.
Here, we employ a capsule network \cite{sabour2017dynamic}, which uses a routing-by-agreement mechanism to capture the correlations among different options and make the final decision. 

Our proposed CEGI framework not only utilizes external commonsense knowledge to generate reasoning evidence but also adopts a capsule network to make the final answer prediction. The explainable evidence and the ablation studies indicate that our method has a large impact on the performance of the commonsense reasoning in reading comprehension. 
The contributions of this paper are summarized as follows: 1) We introduce two evidence generators which are learned from textual and factual knowledge source; 2) We provide an injection method that can infuse both evidences into the contextual reasoning model; 3) We adapt a capsule network to our reasoning model to capture interactions among candidate options when making a decision; 4) We show our  \textbf{CEGI} model outperforms current state-of-the-art models on the CosmosQA dataset and generates richer interpretive evidence which helps the commonsense reasoning.



\section{Related Work}
\subsection{Multi-choice Reading Comprehension}
To model the relation and alignment between the pairs of paragraph, question and option set, various approaches seek to use attention and pursue deep representation for prediction. \citet{tang2019multi} and \citet{wang2018co} model the semantic relationships among paragraph, question and candidate options from multiple aspects of matching. \citet{zhu2018hierarchical} propose a hierarchical attention flow model, which leverages candidate options to capture the interactions among paragraph, question and candidate options. \citet{chen2019convolutional} merge various attentions to fully extract the mutual information among the paragraph, question and options and form the enriched representations. 

\subsection{Commonsense Knowledge Injection}
To empower the model with human commonsense reasoning, various approaches have been proposed on the context-free commonsense reasoning task. The majority of the approaches are focusing on finding the question entity and a reasoning path on the knowledge graph to obtain the answer entity \cite{huang2019cosmos, zellers2018swag, talmor2018commonsenseqa}. For an instance, \citet{lin2019kagnet} construct graphs to represent relevant commonsense knowledge, and then calculate the plausibility score of the path between the question and answer entity. \citet{lv2019graph} extract evidence from both structured knowledge base and unstructured texts to build a relational graph and utilize graph attention to aggregate graph representations to make final predictions. However for contextual commonsense reasoning, it's hard to find a single most relevant entity from the paragraph or question to obtain the correct answer. 

Other approaches focus on enhancing the pre-trained language models through injecting external knowledge into the model and updating the model parameters in multi-task learning \cite{zhang2019ernie, lauscher2019informing, levine2019sensebert}. A knowledge graph injected ERNIE model is introduced in \cite{zhang2019ernie} and a weakly supervised knowledge-pretrained language model (WkLM) is introduced in \cite{xiong2019pretrained}.
They both inject the knowledge through aligning the source with the fact triplets in WikiData. However, the parameters need to be retrained when injecting new knowledge, which could lead to the catastrophic forgetting \cite{mccloskey1989catastrophic}.




\begin{figure*}[t]
\centering
\includegraphics[width=0.9\linewidth]{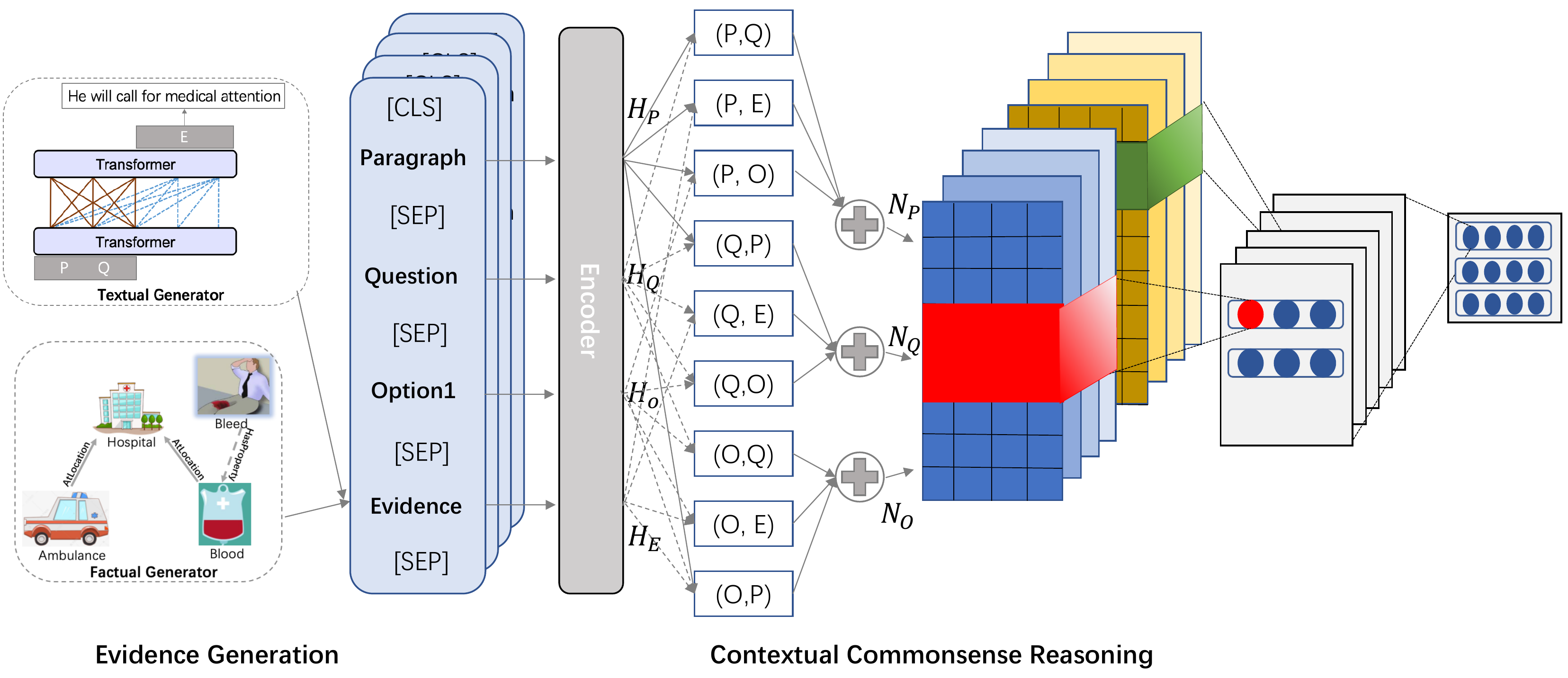}
\vspace{-1em}
\caption{The proposed commonsense evidence generation and injection (\textbf{CEGI}) framework.}
\label{fig:Method}
\vspace{-1em}
\end{figure*}

\section{Task Definition}
In multi-choice reading comprehension, we are given a paragraph $\textbf{P}$ with $t$ tokens $\textbf{P}=[p_{1}, p_{2},\ldots, p_{t}]$, a question $\textbf{Q}$ containing $n$ tokens $\textbf{Q}=[q_{1}, q_{2}, \ldots, q_{n}]$ and the option set with $m$ candidate options $\textbf{O} = \{\textbf{O}_{1}, \textbf{O}_{2}, \ldots, \textbf{O}_{m}\}$, where each candidate option is a text with $h$ tokens $\textbf{O}_i=[o_1, o_2, \ldots, o_h]$, the goal is to select the correct answer $\textbf{A}$ from the candidate option set. For simplicity, we denote $\mathcal{X}=\{\textbf{P},\textbf{Q},\textbf{O}\}$ as one data sample and denote $\textbf{y}=[y_1, y_2,\ldots, y_m]$ as the label, where each $y_i=\textbf{1}(\textbf{O}_{i}=\textbf{A})$ is an indicator function. In the training, $N$ set of $(\mathcal{X}, \textbf{y})^N$ are given, the goal is to learn a model $\mathit{f}: \mathcal{X} \rightarrow \textbf{y}$. In the testing, we need to predict $\textbf{y}^{\rm test}$ given test samples $\mathcal{X}^{\rm test}$. 

When answering a question according to the paragraph, we observe that the context itself often does not provide enough clues to guide us to the correct answer. To this end, we need to know comprehensive information beyond the context and perform commonsense reasoning to address the problem. Hence, we split the task into two parts: evidence generation and answer prediction, respectively. In this paper, our proposed \textbf{CEGI} model addresses both parts accordingly. In the first part, we propose two generators: textual evidence generator and factual evidence generator. In textual evidence generator, our goal is to generate relevant evidence text $\textbf{E}=[e_1, e_2, \ldots, e_{k}]$ given question $\textbf{Q}$ and paragraph $\textbf{P}$. Note that the number of evidence tokens $k$ may vary in different question and paragraph pair. In factual evidence generator, the goal is to generate relevant text that describes the relations between facts where the facts are the entities from paragraph, question and options. In the second part, we aim to learn a classifier $P(\bf{y}|\textbf{P}, \textbf{Q},\textbf{O},\textbf{E})$ that predicts the correct option when a new data sample is given. By using the evidence generated from the first part, we expect the reasoning model can be enhanced with the auxiliary information, especially for those questions that require contextual commonsense reasoning.

\section{Methodology}
To tackle the reading comprehension task with commonsense reasoning, we introduce a commonsense evidence generation and injection (\textbf{CEGI}) framework. The system diagram of the CEGI framework is shown in Fig. \ref{fig:Method}. First, the evidence generation module produces textual evidence and factual evidence. Those generated evidences will be used as auxiliary inputs for the reasoning model. Second, the contextual commonsense reasoning module generates deep contextual features for the paragraph, question, option and evidence. Meanwhile, a bidirectional attention mechanism is applied to the features to capture matching representations of the pair of paragraph, question, option set and evidence. At last, all pairs are concatenated and fed into a convolutional neural network for extracting different linguistic units of the options. The capsule network is then applied to dynamically update the representation vector of the candidate options. The final answer is one of the options with the largest vector norm. We describe more details of each component in the following subsections.

\subsection{Evidence Generation} \label{sec:1}
It is worthy to mention that many commonsense reasoning types, such as causes of events and effects of events, are important factors of understanding the context in reading comprehension. While those factors are often not explicit or given in the paragraph and option set, answering questions regarding commonsense reasoning may become difficult. Hence, to address such a problem, we seek to learn relevant evidence that contains commonsense knowledge. Specifically, we leverage pretrained language models to learn from both context and knowledge graph that may contain reasoning relations. We exploit two kinds of generators, textual evidence generator and factual evidence generator.

\subsubsection{Textual Evidence Generator}
We observe that daily life events often follow a common routine such that when one event happened, the resulting event or the cause of such an event follows a specific pattern. For an example, in Figure~\ref{fig:Intro1}, the given paragraph describes a scenario that the old man is hurt and the young man is making a phone call. If we know that he is calling for medical attention, answering the question would become easy. Hence, the goal of our proposed textual evidence generator is to generate the text that follows daily life event routines. We rely on a pretrained language model to acquire the textual evidence by using GPT2 \cite{radford2018improving} and Uniml \cite{dong2019unified}. Specifically, in the training, we concatenate the paragraph, question and the correct answer as the input to the standard language model \cite{liu2018generating}. 
Accordingly, the textual evidence is the next sentence produced by the language model that we obtained from the training.
Note that we stack $\rm[\textbf{P}$ $\rm [SEP]$ $\textbf{Q}$ $\rm [SEP]$ $\textbf{A}]$ as the input to train the language model. Formally, let $\rm [w^{1}, \ldots, w^{\rm T}]=\rm[\textbf{P}$ $\rm [SEP]$ $\textbf{Q}$ $\rm [SEP]$ $\textbf{A}]$,
the language generation model aims to maximize the following likelihood~\cite{radford2018improving}:
\begin{align}
 \mathcal{L}_{gen} = \sum\nolimits_{i=1}^{T} p(\rm w^{i}|w^{1}, ..., w^{i-1}),
 \label{gen}
\end{align}
where the conditional probability $p(\rm w^{i}|w^{1},\ldots,$ $\rm w^{i-1})=f(\rm w^{1},\ldots, w^{i-1})$ and $\rm f$ is a sequence of operations that (i) converts each token $\rm{w^i}$ into token embedding $\textbf{W}^i_e$ and position embedding $\textbf{W}^i_p$; (ii) transforms them into features with $L$ layers where each layer feature is $\rm \textbf{H}^l({w^i})=h^l(g(\textbf{W}^{i-1}_e,\textbf{W}^{i-1}_p), \textbf{H}^l(\rm{w^{i-1}}))$, and (iii) converts the feature into a probability using a linear classifier by predicting the next token $\rm w^{i}$. 

Moreover, we aim to generate evidence that can discriminate the correct answer from option distractors. Hence, we add a cross-entropy loss into the objective to fine-tune the language model. The text input for the $j$th option is $\textbf{x}_j=\rm[\textbf{P}$ $\rm [SEP]$ $\textbf{Q}$ $\rm [SEP]$ $\textbf{O}_j]$. We use all $N$ samples to 
optimize the following objective (with a regularization term $\lambda$):
\begin{align}
    &\mathcal{L}_{class} = \sum_{(x,y)\in \{\mathcal{X}, \textbf{y}\} } \log (\rm Softmax(\textbf{H}^{L}(w^0) W_{y})),  \\
    &\mathcal{L}_{total} = \mathcal{L}_{gen} + \lambda * \mathcal{L}_{class},
    \label{all_loss}
\end{align}
where $\textbf{H}^{L}(w^0)$ is the last layer feature of the first token and $\rm W_{y}$ is the parameters to learn to predict label $y$.
In the test stage, we only use $\rm [\textbf{P}$ $\rm[SEP]$ $\textbf{Q}]$ as the input to the language model and use the model to generate the next sentence as an evidence. 


\subsubsection{Factual Evidence Generator}
Aside from the textual evidence that contains information about the facts of daily life routine, relations between the facts are also important for question answering. In this work, we propose to utilize a factual knowledge graph to extract facts and relations and use them as additional evidence. Specifically, we use the ConceptNet \cite{speer2017conceptnet}\footnote{ConceptNet is a knowledge graph, which consists of triples obtained from the Open Mind Common Sense entries.} as the base model. We use a knowledge graph completion algorithm \citet{bosselut2019comet} to find new relations to further improve the quality of the generated factual evidence. 

We define $X^{s} = \{x_{0}^{s},..., x_{|s|}^{s}\}$ as subject, $X^{r} = \{x_{0}^{r},..., x_{|r|}^{r}\}$ as relation, and $X^{o} = \{x_{0}^{o},..., x_{|o|}^{o}\}$ as object. We use the $[X^{s}$ $\rm[SEP]$ $X^{r}$ $\rm[SEP]$ $X^{o}]$ triplets as the input to the knowledge graph completion language model in \citet{bosselut2019comet} to generate additional triplets that contain new subject and object relations.
To generate factual evidence, we first extract entities from the given data $\mathcal{X}$. 
We then select the related entities that match the subject $X^s$ in forms of subject-relation-object triplets. After that, we filter the triplets by selecting the subject $X^{s*}$ that follows: 
(i) part-of-speech (POS) tag of $X^{s*}$ word matches the POS tag of the entity word; 
(ii) subject $X^{s*}$ word frequency is less than the word frequency of the object $X^{o}$ plus a threshold $K^o$; 
(iii) subject $X^{s*}$ word is not in the top-$K$ frequent words based on the word frequency table\footnote{\url{https://www.wordfrequency.info/free.asp}}; and 
(iv) the relation $X^r$ in the $(X^{s*}, X^r, X^o)$ triplets connects no more than $K^r$ objects from the same subject $X^{s*}$. $K$, $K^o$ and $K^r$ are the hyper-parameters. 
Finally, we convert the filtered triplets into a nature language sequences as our factual evidences. For example, “(trouble, Partof, life)” would be converted to “trouble is part of life”. 


\subsection{Model Learning with Contextual Commonsense Reasoning} \label{sec:2}
After the relevant reasoning evidences are generated, the goal is to combine the evidence with the given data and then build a reasoning model to make a selection for the correct answer. 
In the following, we introduce our proposed contextual commonsense reasoning module, which utilizes contextual encoding, evidence injection and a capsule network components to make the prediction. 

\textbf{Contextual Encoding}
Recently, RoBERTa \cite{liu2019roberta} has shown to be effective and powerful in many natural language processing tasks and it is potentially beneficial for generating deep contextual features as well. Here, we use RoBERTa as an intermediate component to generate hidden representation of paragraph, question, the $i$th option and evidence $[\textbf{H}^i_{\rm cls}, \textbf{H}^i_{\rm P},\textbf{H}^i_{\rm sep} \textbf{H}^i_{\rm Q}, \textbf{H}^i_{\rm sep}$ $\textbf{H}^i_{\rm O_i}, \textbf{H}^i_{\rm sep},\textbf{H}^i_{\rm E}] = 
\rm Encode([\rm [CLS],$ $\textbf{P},$ $\rm [SEP],$ $\textbf{Q},$ $\rm [SEP],$ $\textbf{O}_i,$ $\rm [SEP], \textbf{E}]) $. 
We use the last layer of the RoBERTa model to encode, and thus the function $\rm Encode(\cdot)$ returns the last layer features for each token. The corresponding features of paragraph, question, option and evidence are $\textbf{H}^i_{\rm P} \in \mathcal{R}^{d\times t}$, $\textbf{H}^i_{\rm Q} \in \mathcal{R}^{d\times n}$, $\textbf{H}^i_{\rm O_i} \in \mathcal{R}^{d\times h}$ and $\rm \textbf{H}^i_{E} \in \mathcal{R}^{d\times k}$, where $d$ is the dimension of the feature. Since we have $m$ options, we have $m$ set of features. 

\textbf{Evidence Injection}
Given the previously generated evidence representation $\textbf{H}^i_{\rm E}$. We aim to integrate it with the paragraph $\textbf{H}^i_{\rm P}$, question $\textbf{H}^i_{\rm Q}$ and option $\textbf{H}^i_{\rm O_i}$. Here, we adopt the attention mechanism used in QANet \cite{yu2018qanet} to model the interaction between $\rm \textbf{H}^i_{E}$ and the paragraph $\textbf{H}^i_{\rm P}$:
\begin{align}
    \rm \textbf{S}_{iP}^{E} &= \rm Att(\textbf{H}^i_{E}, \textbf{H}^i_{P})=Softmax(\textbf{H}^{iT}_{P}\rm W_{g}\textbf{H}^i_{E}) \\
    \rm \textbf{G}_{iP}^{E} &= \rm \textbf{H}^i_{E} \textbf{S}_{iP}^{E^T} ,
\end{align}
where $\rm W_{g} \in \mathcal{R}^{d \times d}$ is the bi-linear model parameter matrix. Since $\rm \textbf{S}_{iP}^{E}\in\mathcal{R}^{t\times k}$ is the activation map (attention weights) between each token in $\textbf{P}$ and each token in $\textbf{E}$, the learned relation representation $\rm \textbf{G}_{iP}^{E}\in\mathcal{R}^{d \times t}$ of the paragraph $\textbf{P}$ contains evidence information $\textbf{E}$. The other two relations $\rm \textbf{G}_{iP}^{Q}$ and $\rm \textbf{G}_{iP}^{O_i}$ regarding $\textbf{P}$ can be generated accordingly.
Similarly, we can model the other interactions for question $\textbf{Q}$ as  $\rm \textbf{G}_{iQ}^{P}, \rm \textbf{G}_{iQ}^{E}, \rm \textbf{G}_{iQ}^{O_i}$, and each option $\textbf{O}_i$ as $\rm \textbf{G}_{iO_i}^{Q}, \rm \textbf{G}_{iO_i}^{E}$ and $\rm \textbf{G}_{iO_i}^{P}$.

To incorporate the relation information, we use the co-matching algorithm introduced in \citet{wang2018co} to generate the final representation of the input. First, we obtain the matching result between the paragraph and the question as follows: 
\begin{align}
 \rm \textbf{M}_{iP}^{Q} = \rm (W_{m}[
 \textbf{G}_{iP}^{Q} \ominus \textbf{H}^i_{P};
 \textbf{G}_{iP}^{Q}  \odot \textbf{H}^i_{P}] + b_{m}\otimes \textbf{1} )^{+}, 
\end{align}
where $(\cdot)^{+}$ denotes ReLU function, $\textbf{1}=[1,1,\ldots,1]^T\in \mathcal{R}^{t\times 1}$ is vector of all ones, and $\rm W_{m} \in \mathcal{R}^{d\times 2d}$ and $\rm b_{m} \in \mathcal{R}^{d\times 1}$ are the model parameters.
Following \citet{tai2015improved} and \citet{wang2018co}, we use notation $\ominus$ and $\odot$ as the element-wise subtraction and multiplication between two matrices and $\otimes$ as outer product of two vectors. 
Similarly, we can obtain the other pairs as $\rm \textbf{M}_{iP}^{E}, \rm \textbf{M}_{iP}^{O_i}, \ldots, \textbf{M}_{iO_i}^{P}$.
In the next step, we concatenate all the pairs regarding $\textbf{P}$ as
\begin{align}
\rm \textbf{C}_{iP} = \begin{bmatrix}
\rm  \textbf{M}_{iP}^{Q} :
\textbf{M}_{iP}^{O_i} :
\textbf{M}_{iP}^{E} \label{end}
\end{bmatrix}\in \mathcal{R}^{3d\times t},
\end{align}
where $[:]$ denotes the vertical concatenation operation. Each column $\textbf{c}_i$ is the co-matching state that concurrently matches a paragraph token with the question, candidate option and the evidence. Accordingly, we can obtain the question representation $\rm \textbf{C}_{iQ}$ and option representation $\rm \textbf{C}_{iO_i}$.
Finally, we concatenate them all to obtain the final representation $\textbf{F} = [\textbf{C}_{1}, \ldots, \textbf{C}_{m}]\in \mathcal{R}^{\rm 3d\times m(t+n+h)}$, where each $\rm \textbf{C}_{i}=\rm [\textbf{C}_{iP}, \textbf{C}_{iQ}, \textbf{C}_{iO_{i}}] \in \mathcal{R}^{3d\times(t+n+h)}$. 

Since the final representation only contains the fine-grid token-level information, we employ a convolutional neural network (CNN) to extract higher level (phrase-level) patterns.  To generate phrase patterns with different size, we use two convolutional kernels: size $1\times 2$ with stride 2 and size $1\times 4$ with stride 4 to convolve with $\textbf{F}$ along the dimension of hidden state. In other words, such an operation extracts non-overlapping moving windows on $\textbf{F}$ with window size $2$ and $4$.
\begin{align}
\nonumber & \textbf{R}_{1} = \rm MaxPooling_{1\times2}\{CNN_{1\times2}(\textbf{F})\} \\
\nonumber & \textbf{R}_{2} = \rm MaxPooling_{1\times1}\{CNN_{1\times4}(\textbf{F})\}
\end{align}
To ensure $\textbf{R}_{1}$ and $\textbf{R}_{2}$ have the same dimension, we use a max pooling of size $1\times 2$ with stride 2 for $\textbf{R}_{1}$ and a max pooling of size $1\times 1$ with stride 1 for $\textbf{R}_{2}$.
We concatenate $\textbf{R}_1$ and $\textbf{R}_2$ to generate phrase-level representation $\textbf{L} = [\textbf{R}_{1}, \textbf{R}_{2}]\in \mathcal{R}^{3d\times m((t+n+h)/2)}$. With $\textbf{L}$, to predict the final answer, one of the commonly applied operation is to simply take the maximum over the hidden dimension of length $(t+n+h)/2$. However, the max operation only consider the most significant phrase for each candidate without aware of the others. To explore the correlation between options and dynamically select the optimal one, we use dynamic routing-by-agreement algorithm represented in \citet{sabour2017dynamic}. Specifically, we convert $\textbf{L}_{i}$ to a capsule $\rm \textbf{v}_j$ using the following steps:
\begin{align}
\nonumber \rm \hat{\textbf{L}}_{\rm j|i} &= W_{ij} \textbf{L}_{i}, ~~
\textbf{s}_{j} = \sum\nolimits_{i=1}^{(t+n+h)/2}c_{ij}\cdot \hat{\textbf{L}}_{j|i},\\
\nonumber \rm \textbf{v}_{j} &= \frac{||\textbf{s}_{j}||^{2}}{1+||\textbf{s}_{j}||^{2}} \frac{\textbf{s}_{j}}{||\textbf{s}_{j}||},
\end{align}
where $\textbf{L}_{i}$ is the $i$th column vector of $\textbf{L}$, 
affine transformation matrix $\rm W_{ij}$ and weighting $c_{ij}$ are the capsule network model parameters. 
The learned $\hat{\textbf{L}}_{\rm j|i}$ denotes the ``vote'' of the capsule $\rm j$ for the input capsule $\rm i$. The agreement of ``prediction vector'' $\hat{\textbf{L}}_{\rm j|i}$ between the current $j$th output and $i$th parent capsule is captured by the coupling coefficients $c_{\rm ij}$. The value of $c_{\rm ij}$ would increase if higher level capsule $\textbf{s}_{j}$ and lower lever capsule $\textbf{L}_{i}$ highly agreed.

\textbf{Model Learning}
If an option $\textbf{O}_j$ is the correct answer, we would like the top-level capsule $\textbf{v}_{j}$ to have a high energy, otherwise, we expect the energy of $\textbf{v}_{j}$ to be low.
Since the $L_2$-norm (square root of the energy) of the capsule vector $\rm \textbf{v}_{j}$ represents the scoring of how likely the $j$th candidate is the correct answer, we use the following loss function \cite{sabour2017dynamic} to learn the model parameters:
\begin{align}
  \nonumber   \mathcal{L}_{pre}=& \sum\nolimits_{j=1}^{m}\{ y_i \cdot \max(0, m^{+} - ||\textbf{v}_{j}||)^{2} \\
  & \hspace{-0.2cm} + \lambda_{1} (1-y_i) \max(0, ||\textbf{v}_{j}|| - m^{-})^{2} \}
\end{align}
where $\lambda_{1}$ is a down-weighting coefficient, $m^{+}$ and $m^{-}$ are margins. 
In our experiments, we set $m^{+}=0.9$, $m^{-}=0.1$, $\lambda_{1}=0.5$.

\section{Experiments}
In the experiment, we evaluate the performance of our proposed CEGI framework from different aspects, including evidence generation tasks and the answer prediction of contextual commonsense reasoning tasks. 

\subsection{Dataset and Baseline}
\textbf{CosmosQA} is the dataset that is designed for reading comprehension with commonsense reasoning \cite{huang2019cosmos}. Samples are collected from people's daily narratives and the type of questions are concerning the causes or effects of events. Particularly answering the questions require contextual commonsense reasoning over the considerably complex, diverse, and long context. In general, the dataset contains a total of 35.2K multiple-choice questions, including 25262 training samples, 2985 development samples, and 6963 testing samples.\footnote{The CosmosQA dataset can be obtained from  \url{https://leaderboard.allenai.org/cosmosqa/}}

\noindent\textbf{Baseline}
We categorize baseline methods into the following three groups: 
$1.$ Co-Matching \cite{wang2018co}, Commonsense-RC \cite{wang2018yuanfudao}, DMCN \cite{zhang2019dcmn}, Multiway \cite{huang2019cosmos}. 
$2.$ GPT2-FT \cite{radford2018improving}, BERT-FT \cite{devlin2018bert}, RoBERTa-FT \cite{liu2019roberta}.
$3.$ Commonsense-KB \cite{li2019teaching}, K-Adapter \cite{wang2020k}.
The baseline details are in appendix A.2.

\begin{table}[!htpb]
\caption{Comparison of approaches on CosmosQA (Accuracy $\%$) from the AI2 Leaderboard. T+F means using generated textual and factual evidence together.}
\vspace{-0.5em}
\centering
\resizebox{0.49\textwidth}{!}{
\begin{tabular}{l >{\columncolor[gray]{0.7}}c>{\columncolor[gray]{0.6}}c}
\toprule[1.5pt]
 Model          & Dev & Test \\ \hline
 Co-Matching \cite{wang2018co}    &  45.9 & 44.7    \\
 Commonsense-RC \cite{wang2018yuanfudao} & 47.6 & 48.2     \\
 DMCN   \cite{zhang2019dcmn}         &  67.1 & 67.6   \\
 Multiway   \cite{huang2019cosmos}    & 68.3 & 68.4     \\\hline
 GPT-FT    \cite{radford2018improving}      & 54.0 & 54.4     \\ 
 BERT-FT   \cite{devlin2018bert}     &  66.2 & 67.1     \\
 RoBERTa-FT  \cite{liu2019roberta}   & 79.4 & 79.2     \\ \hline 
 Commonsense-KB \cite{li2019teaching} & 59.7  &  $\backslash$     \\
 K-Adapter  \cite{wang2020k}    &  81.8 & $\backslash$   \\ \hline
CEGI(T+F)       & \textbf{83.8} & \textbf{83.6}      \\ \hline
 Human          &    $\backslash$  &   94.0 \\  
\toprule[1.5pt]
\end{tabular}
}
\label{acc_cosmos}
\vspace{-1em}
\end{table}

\subsection{Experimental Results and Analysis}
Table \ref{acc_cosmos} shows the performance of different approaches reported on the AI2 Leaderboard.\footnote{\url{https://leaderboard.allenai.org/cosmosqa/}}. Comparing to all methods, our proposed model CEGI(T+F) has the highest accuracy on both development set and test set. 
Most of the reading comprehension approaches utilize the attention mechanism to capture the correlations between paragraph, question and option set, therefore, the model tends to select the one option that is semantically closest to the paragraph. Among all of the group 1 methods, Multiway has the highest accuracy of $68.3\%$.
Group 2 methods consider deep contextual representation of the given paragraph, question and option set, and increase the performance.
Comparing group 2 methods with group 1 methods, RoBERTa-FT, which uses dynamic masking and large mini-batches strategy to train BERT, gains $11.1\%$ accuracy increase compared to Multiway.

However, it is worthy to mention that more than $83\%$ of correct answers are not in the given passages in the CosmosQA dataset. Hence, multi-choice reading comprehension models do not gain big improvement as they tend to select the choice which has the most overlapped words with the paragraph without commonsense reasoning. Even though, group 2 methods consider connecting the paragraph with question and option through a deep bi-directional strategy, the reasoning for question answering is still not well-addressed in the models.
By utilizing additional knowledge, Commonsense-KB or K-Adapter teach pretrained models with commonsense reasoning. K-Adapter gains $2.4\%$ accuracy increase than RoBERTa-FT. Those approaches leverage the structured knowledge but fail to produce a prominent prediction improvement.
Comparing our CEGI approach with RoBERTa, we gain a $4\%$ increase and $2\%$ increase than K-Adapter, which demonstrates that injecting evidence is beneficial and incorporating interactive attentions can further enhance the model. 

\subsection{Evidence Evaluation}
In this section, we investigate the generated evidence from the textual generator and factual generator. Moreover, we study the quality of the generated evidence on another dataset---CommonsenseQA.

\subsubsection{Textual Evidence Generator} 
\textbf{Dataset} Open Mind Common Sense (OMCS) corpus \cite{singh2002open} is a crowd-sourced knowledge database of commonsense statements \footnote{\url{https://github.com/commonsense/conceptnet5/wiki/Downloads}}, where its English dataset contains a million sentences from over 15,000 contributors. We consider using this dataset to pretrain the textual evidence generator and using CosmosQA to fine-tune the generator.

\noindent\textbf{Setup} 
We use both BERT and GPT2 model to generate evidence and compare the results. To obtain a language model that contains representation of facts, we first pretrain both models with the OMCS data using the loss function in Eq. \ref{gen}. Then we use CosmosQA data to fine-tune the pretrained model using multi-task loss in \ref{all_loss}.

\noindent\textbf{Metrics} In line with prior work \cite{wang2019bert}, we evaluate the performance of evidence generation based on quality and diversity. In terms of quality, we follow \citet{yu2017seqgan} and compute the BLEU score between the generated evidence and the ground truth evidence to measure the similarity. The perplexity (PPL) score is also reported as a proxy for fluency. 
In terms diversity, we consider using self-BLEU \cite{zhu2018texygen}, which measures how similar between two generated sentences. Generally, a higher self-BLEU score implies that the model has a lower diversity.

\noindent\textbf{Results} From Table \ref{generation}, we observe that, compared to CEGI-GPT2, the CEGI-BERT generator has higher diversity (Self-BLEU decreases $4$ for bi-gram and decreases $2.1$ for tri-gram) but lower quality (BLEU decreases $1.3$ for tri-gram and PPL increases $27.1$). Even though the perplexity on CEGI-BERT is as good as CEGI-GPT2, after reading the samples, we find out that many of the generated language are fairly coherent. 
For a more rigorous measure of generation quality, we collect human judgments on sentences for 100 samples using a four-point scale (the higher the better). For each sample, we ask three annotators to rate the sentence on its fluency and take the average of the three judgments as the sentence’s fluency score. For CEGI-BERT and CEGI-GPT2, we get mean scores of 3.21, 3.17 respectively. Those results imply that generated evidence are semantically consistent with the correct evidence and can be used as auxiliary knowledge for the reasoning step. 
 
\begin{table}[!htb]
\caption{Generation performance on CosmosQA.}
\centering
\resizebox{0.45\textwidth}{!}{
\begin{tabular}{llllll}
\toprule[1.5pt]
            & \multicolumn{3}{c}{Quality}                            & \multicolumn{2}{c}{Diversity}  \\ \cline{2-6} 
Model       & \multicolumn{2}{c}{BLEU($\uparrow$)} & PPL($\downarrow$) & \multicolumn{2}{c}{Self-BLEU($\downarrow$)} \\ \cline{2-3} \cline{5-6}  
            & n=2 & n=3 &     & n=2       & n=3           \\ 
\hline
CEGI-BERT &  \textbf{40.8} &  32.2   &    153.8     & \textbf{30.5}  &    \textbf{14.7}             \\
CEGI-GPT2    &  39.8 & \textbf{33.5}    &    \textbf{126.7}     & 34.2   &    16.6           \\
\bottomrule[1.5pt]
\end{tabular}
}
\label{generation}
\vspace{-1em}
\end{table}

\begin{table}[ht]
\caption{Generation performance on ConceptNet}
\centering
\resizebox{0.4\textwidth}{!}{
\begin{tabular}{lllll}
\toprule[1.5pt]
Model  & PPL & Score & N/T sro & N/T o \\ \hline
LSTM-s & $\backslash$  & 60.83 & \textbf{86.25}   & 7.83  \\
CKBG   & $\backslash$   & 57.17 & \textbf{86.25}   & \textbf{8.67}  \\ \hline
CEGI-BERT  & 4.89    &  92.19   &   65.32    &    4.12    \\
CEGI-GPT2  & \textbf{4.58}    &  \textbf{93.89}   &   61.72    &    3.90        \\
\bottomrule[1.5pt]
\end{tabular}
}
\label{concept}
\vspace{-1em}
\end{table}

\subsubsection{Factual Evidence Generator} 
\textbf{Dataset} ConceptNet\footnote{\url{https://ttic.uchicago.edu/~kgimpel/commonsense.html}} is a commonsense knowledgebase of the most basic things a person knows. We use the 100K version of the training set in ConceptNet, which contains 34 relation types, to train the factual evidence generator. Tuples within the data are in the standard $<s, r, o>$ form.

\noindent\textbf{Setup} 
We set $s$ and $r$ as input for both GPT2 and BERT and use them to generate the new object $o$. To compare with our proposed GPT2 model and BERT model, we include a LSTM model (LSTM-s) and the BiLSTM model (CKBG) in \cite{saito2018commonsense}. We train the LSTM model to generate $o$, and we train the CKBG model from both directions: $s$, $r$ as input and $o$, $r$ as input. 

\noindent\textbf{Metrics} Similar to the textual evidence generation task, we use PPL to evaluate our model on relation generation. To evaluate the quality of generated knowledge, we also report the number of generated positive examples that are scored by the Bilinear AVG model \cite{li2016commonsense}. ``N/T sro'' and ``N/T o'' are the proportions of generated tuples and generated objects which are not in the training set.  

\noindent\textbf{Results}
As we observed from Table \ref{concept}, CEGI-GPT2 has the lowest PPL ($4.58$) and highest score ($93.89$), which indicates the CEGI-GPT2 model is confident and accurate at the generated relations. Even though the generated tuples on LSTM-s and CKGB model has high  ``N/T sro'' (both are $86.25\%$) and ``N/T o'' ($7.83\%$ and $8.67\%$ respectively), which means they generate novel relations and expand the knowledge graph, the generated nodes and relations may not be correct. We still need to rely on the Score to evaluate and they do poorly ($60.83\%$ and $57.17\%$ respectively) in terms of Score. Since our proposed CEGI-GPT2 and CEGI-BERT model have high Score and low PPL, we believe that both models can produce high-quality knowledge and still be able to extend the size of the knowledge graph. 

\subsubsection{Evidence Evaluation on CommonsenseQA}
\textbf{CommonsenseQA}\footnote{\url{https://www.tau-nlp.org/commonsenseqa}} is a multi-choice question answering dataset, which contains roughly $12K$ questions with one correct answer and four distractor answers. Since the CommonsenseQA data only requires different types of commonsense knowledge to predict the correct answers, it does not contain paragraphs compared to CosmosQA. 
We use our textual generator and factual generator to generate evidence using CommonsenseQA data and use that to test the performance on answer prediction. 
To train our proposed textual evidence generator, we use Cos-e\footnote{\url{https://github.com/salesforce/cos-e}} as the ground truth evidence. Cos-e uses Amazon Mechanical Turk to provide reasoning explanations for the CommonsenseQA dataset. To train our proposed factual evidence generator, we follow the same procedure as described in subsection 4.1.2. To predict the answer based on both evidence, we prepare the input as $[\textbf{Q}$ $\rm [SEP]$ $\textbf{O}_{i}$ $\rm [SEP]$ $\textbf{E}$ $]$ to the RoBERTa model.  

\noindent\textbf{Baselines} KagNet \cite{lin2019kagnet}, Cos-E \cite{rajani2019explain},
DREAM \cite{lv2019graph}, RoBERTa + KE, RoBERTa + IR and RoBERTa + CSPT \cite{lv2019graph}. All baselines use extracted knowledge from ConceptNet or Wikipedia. The details are in the appendix A.2. 
\begin{table}[ht]
\caption{Accuracy ($\%$) of different models on CommonsenseQA development set}
\centering
\resizebox{0.35\textwidth}{!}{
\begin{tabular}{ll}
\toprule[1.5pt]
Model       & Acc  \\ \hline
KagNet \cite{lin2019kagnet}     & 62.4 \\
Cos-E    \cite{rajani2019explain}   & 64.7 \\
DREAM    \cite{lv2019graph}   & 73.0 \\
RoBERTa+CSPT \cite{lv2019graph}& 76.2 \\
RoBERTa+KE  \cite{lv2019graph} & 77.5 \\
RoBERTa+IR  \cite{lv2019graph} & 78.9 \\ \hline
RoBERTa + T  & 78.8 \\
RoBERTa + F  & 77.6 \\
RoBERTa + (T+F)  &\textbf{79.1} \\
\bottomrule[1.5pt]
\end{tabular}
}
\label{commonsenseqa}
\vspace{-1em}
\end{table}

\noindent\textbf{Result}
Results on CommonsenseQA datasets are summarized in Table \ref{commonsenseqa}. RoBERTa + T, RoBERTa + F and RoBERTa + (T+F) includes textual evidence, factual evidence and both evidence together respectively. We observe that our model RoBERTa + T  and RoBERTa + F can produce competitive performance compared to all baselines. By utilizing both textual knowledge and factual knowledge, our approach outperforms RoBERTa+IR and achieves the highest accuracy $79.1\%$.

\subsection{Ablation Study}
To evaluate the contributions of individual components of our proposed framework, we use an ablation study. Table \ref{ablation} summarizes ablation studies on the development set of CosmosQA from several aspects: the influence of the generated evidence; which evidence is better, textual or factual; the influence of the capsule network. In the evidence row, we keep using the capsule network but with or without knowledge injection. In the capsule row, we keep the textual and factual evidences and test on using and not using a capsule network.
From the result, we can see that injecting generated explainable evidence can help the model achieve a better performance in terms of accuracy. Using generated textual evidence and factual evidence together can benefit more. Using capsule network significantly improves the reasoning performance, we doubt that is due to the hierarchical structure information from both token-level and phrase-level are extracted by Capsule network.

\begin{table}[ht]
\caption{Accuracy ($\%$) of different models on Cosmos development set}
\centering
\resizebox{0.35\textwidth}{!}{
\begin{tabular}{ccc}
\cline{2-2}
\toprule[1.5pt]
~~~~~~~ Model                   & Acc \\ \hline
Evidence    & without               & 82.1      \\
            & + Text-injection       &    82.4       \\
            & + KG-injection       &      82.5      \\ 
            & + Text+KG-injection    &       83.2    \\ \hline
Capsule     & without       &      81.8      \\  
            & + capsule         &     83.1   \\ 
\bottomrule[1.5pt]
\end{tabular}
}
\label{ablation}
\vspace{-1em}
\end{table}




\section{Conclusion}
In this paper, we proposed a commonsense evidence generation and injection model to tackle reading comprehension. Both textual and factual evidence generators were used to enhance the model for answering questions which requires commonsense reasoning.
After the evidences are generated, we adopted attention mechanism to find the relation and match between paragraph, question, option and evidence. We have used convolutional network to capture the multi-grained features. To capture diverse features and iteratively make a decision, we proposed using a capsule network that dynamically capture different features to predict the answer. 
The AI2 Leaderboard of CosmosQA task demonstrated that our method can tackle commonsense-based reading comprehension pretty well and it outperformed the current state-of-the-art approach K-Adapter with a $2\%$ increase in term of accuracy. 
Experiments regarding the evidence generators showed that the generated evidence is human-readable and those evidences are helpful for the reasoning task.


\section{Acknowledge}
This work is supported in part by NSF under grants III-1526499, III-1763325, III-1909323, and CNS-1930941. 


\clearpage

\begin{appendices}
\section{Appendix}
\subsection{Training Details}
In CosmosQA experiments, we use pretrained weight of RoBERTa$\_$large. We set the routing iterations of capsule network as 3. We run experiments on a 24G Titan RTX for 5 epochs, set the max sequence length to 256.
For hyper-parameters, batch size is chosen from $\{$8, 16, 24, 32$\}$, learning rate is chosen from $\{$2e-5, 1e-5, 5e-6$\}$ and warmup proportion is chosen from $\{$0, 0.1, 0.2, 0.5$\}$. For CEGI(F+L), the best performance is achieved at batch size=24, lr=1e-5, $\rm warmup\_proportion$=0.1 with 16-bit float precision. 
GPT2 with 12-layer and BERT$\_$base model are used in evidence generation. In textual evidence generation, we set $\lambda$ in Eq. \ref{all_loss} to $0.5$, max sequence length to 40, batch size to 32 and the learning rate to 6.25e-0.5.
In factual evidence generation, we set max sequence length to 15, batch size to 64, the learning rate to 1e-5. For both generators, we train 100000 iterations with early stop.

\subsection{Baseline Methods}
\textbf{Cosmos Baselines}
\\\noindent$1.$ \textbf{Co-Matching} \cite{wang2018co} captures the interactions between paragraph with question and option set through attention. 
\textbf{Commonsense-RC} \cite{wang2018yuanfudao} performs three-way unidirectional attention to model interactions between paragraph, question, and option set. 
\textbf{DMCN} \cite{zhang2019dcmn} applies dual attention between paragraph and question or option set using BERT encoding output. 
\textbf{Multiway} \cite{huang2019cosmos} uses BERT to learn the semantic representation and uses multiway bidirectional interaction between each pair of input paragraph, question and option set. 
\\\noindent$2.$ \textbf{GPT2-FT} \cite{radford2018improving}, \textbf{BERT-FT} \cite{devlin2018bert} and \textbf{RoBERTa-FT} \cite{liu2019roberta} are the pretrained transformer language models with additional fine-tuning steps on CosmosQA. 
\\\noindent$3.$ \textbf{Commonsense-KB} \cite{li2019teaching} uses logic relations from a commonsense knowledge base (e.g., ConceptNet\footnote{\url{http://conceptnet.io/}}) with rule-based method to generate multiple-choice questions as additional training data to fine-tune the pretrained BERT model.
\textbf{K-Adapter} \cite{wang2020k} infuses commonsense knowledge into a large pre-trained network. \\
\textbf{CommonsenseQA Baselines} \\
\textbf{KagNet} \cite{lin2019kagnet} uses ConceptNet as extra knowledge and proposes a knowledge-aware graph network and finally scores answers with graph representations.
\textbf{Cos-E} \cite{rajani2019explain} constructs human-annotated evidence for each question and generates evidence for test data.
\textbf{DREAM} \cite{lv2019graph} adopts XLNet-large as the baseline and extracts evidence from Wikipedia. \textbf{RoBERTa + KE}, \textbf{RoBERTa + IR} and \textbf{RoBERTa + CSPT} \cite{lv2019graph} adopt RoBERTa as the baseline and utilize the evidence from Wikipedia, search engine and OMCS, respectively.

\subsection{Case Study}
To verify the generated evidence performance, we perform case studies on textual generator and factual generator. In addition, we also show a case that the proposed capsule network can help to select the answer by comparing with the other options. 

\begin{figure}[htb]
\centering
\includegraphics[width=0.95\linewidth]{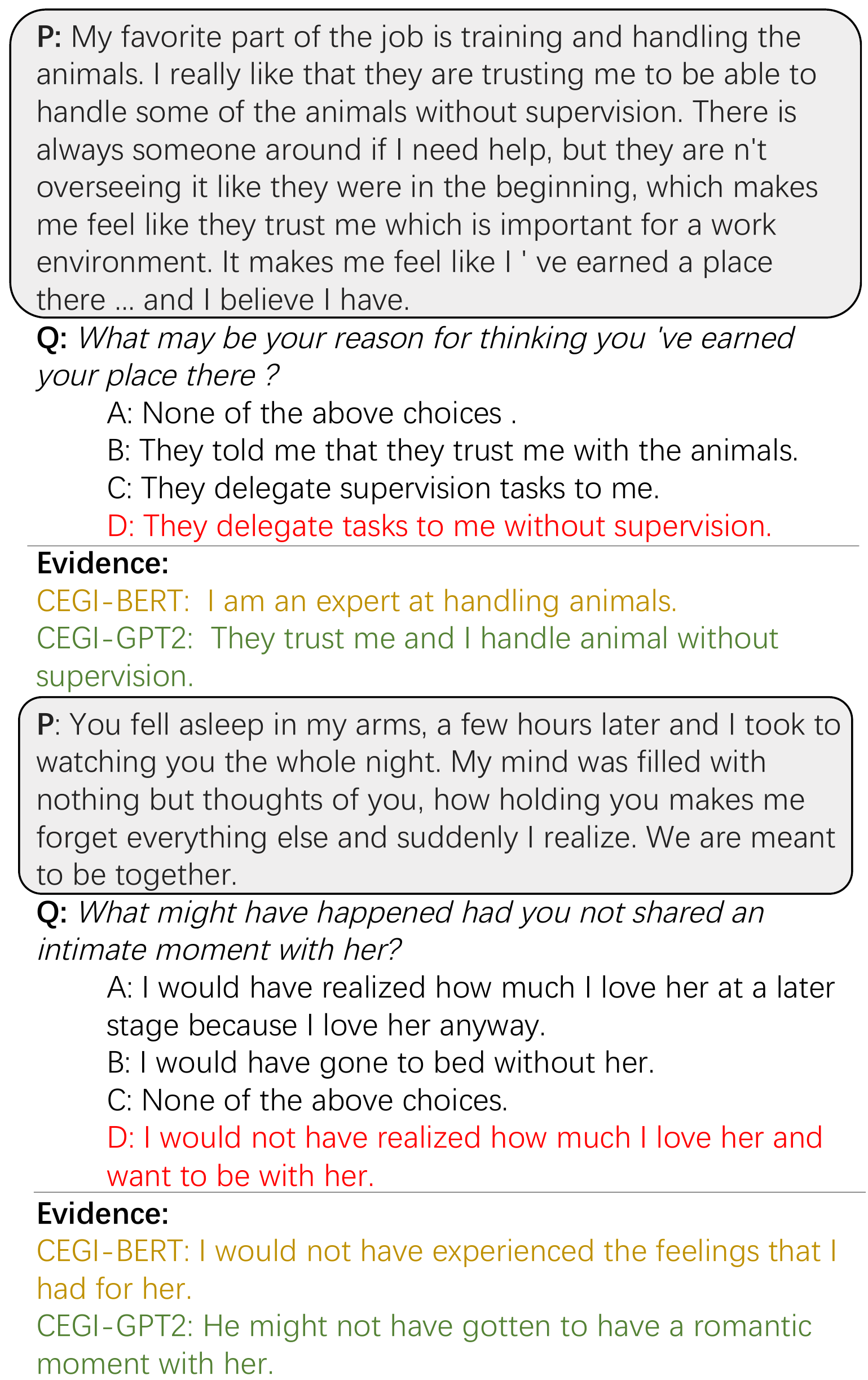}
\caption{Examples of textual evidence generator.}
\label{fig:exp1}
\vspace{-1em}
\end{figure}

\textbf{Case Study on Textual Generator}
We show examples of automatically generated evidences by CEGI-GPT2 and CEGI-BERT in Figure \ref{fig:exp1}. We observe that using the multi-tasking loss, CEGI-BERT and CEGI-GPT2 generates more accurate evidence. Moreover, using those generated evidences is helpful for predicting the correct answer. In the first example, the evidence generated by CEGI-GPT2 ``They trust me and I handle animal without supervision.'' can help select the Answer D ``They delegate tasks to me without supervision.''
In the second example, the evidence generated by CEGI-BERT ``I would not have experienced the feelings that I had for her.'' is close to the Answer D.
\begin{figure}[htb]
\centering
\includegraphics[width=0.95\linewidth]{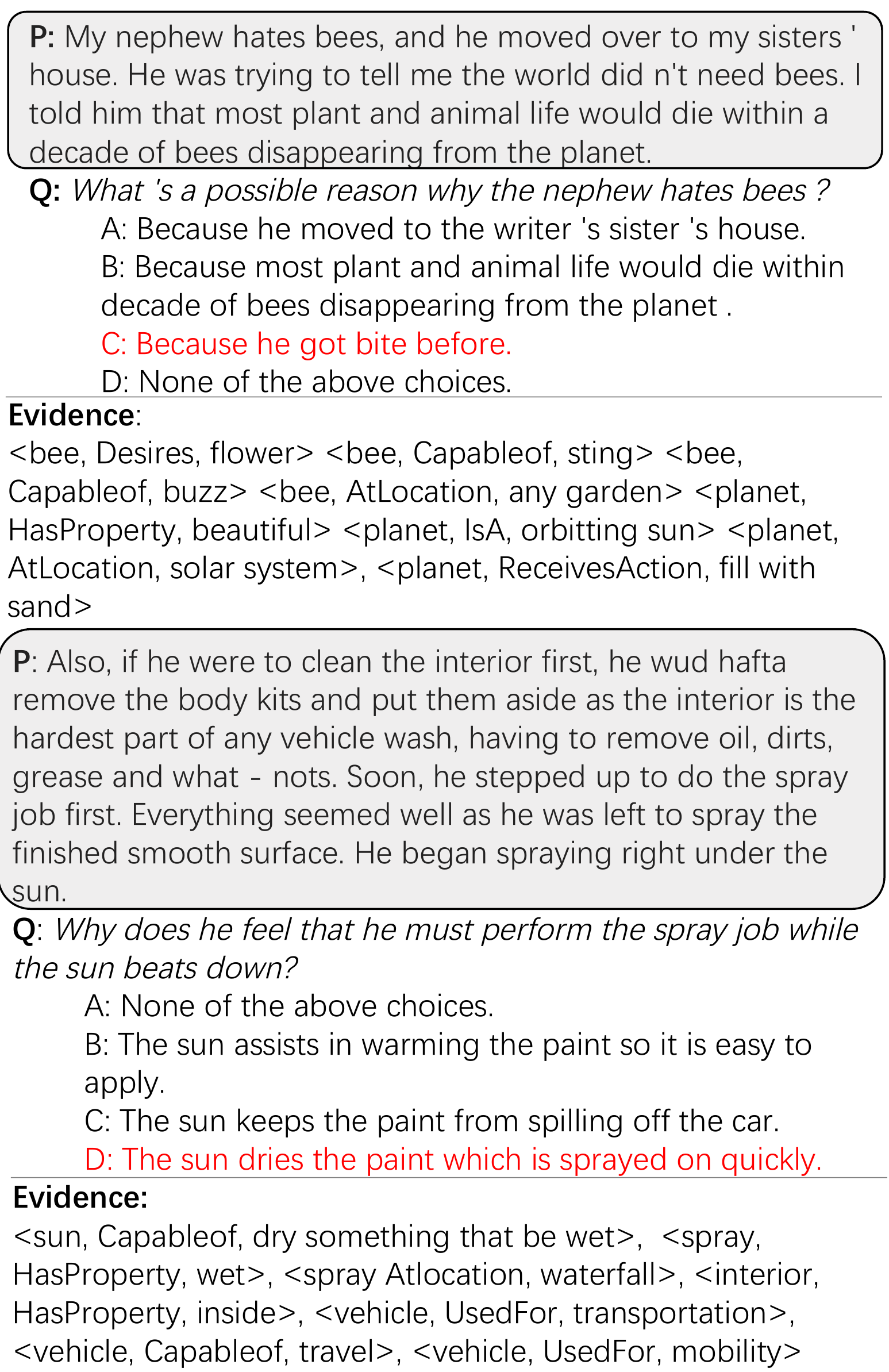}
\caption{Examples of factual evidence generator.}
\label{fig:exp2}
\vspace{-1em}
\end{figure}

\textbf{Case Study on Factual Generator}
Figure \ref{fig:exp2} shows the examples of evidences generated by the factual generator. In the first example, from evidence, we know ``bee is capable of sting'', so option C ``Because he got bite before'' will be the correct answer. Some options like B ``Because most plant and animal life would die within decade of bees disappearing from the planet'' appear in the context ``I told him that most plant and animal life would be die within a decade of bees disappearing from the planet'', and thus without the evidence it could puzzle the model to select B. In the second example, we have the evidence ``sun has capable of drying something that be wet'' and ``spray has property wet'', so it is easy to reach the correct answer D ``The sun dries the paint which is sprayed on quickly''. 

\begin{figure}[htb]
\centering
\includegraphics[width=0.95\linewidth]{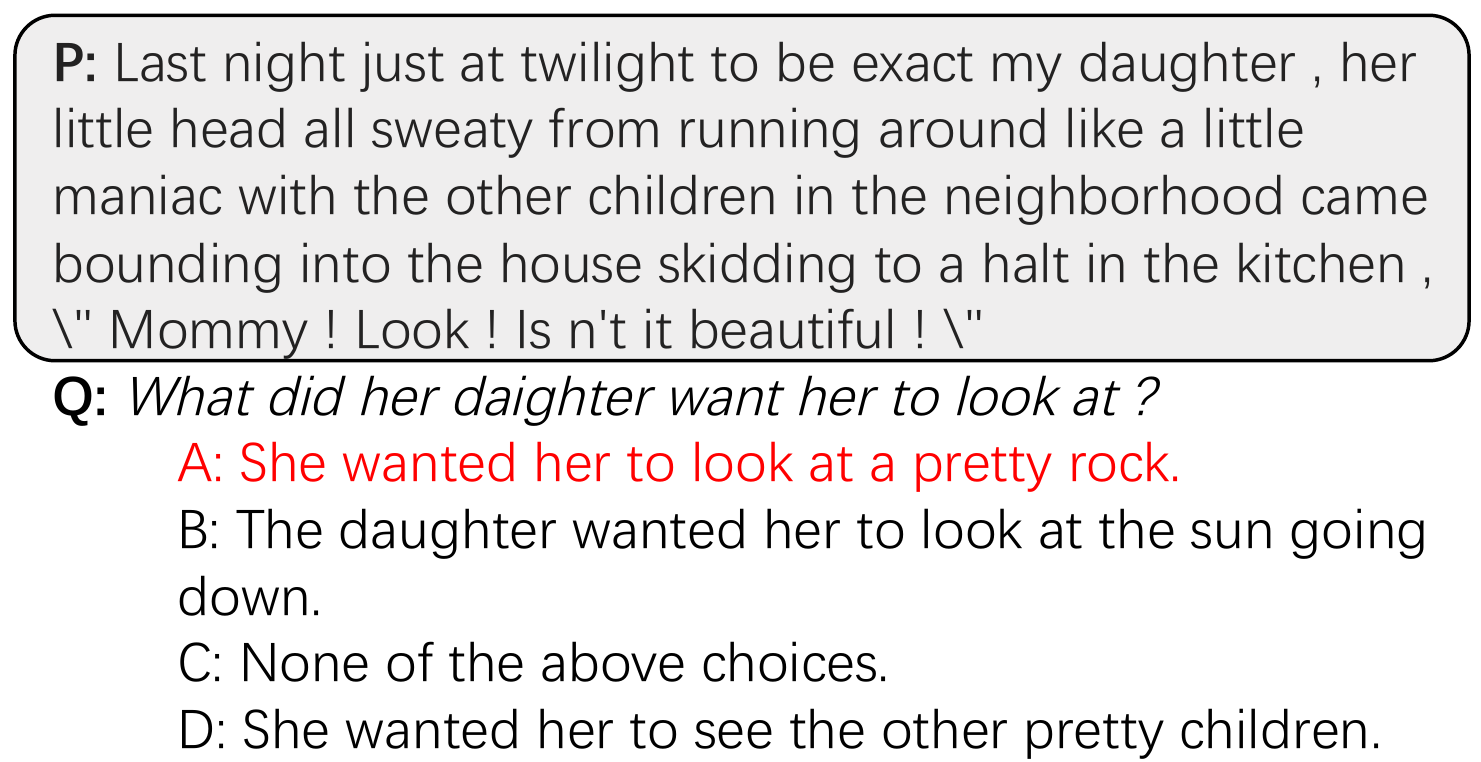}
\caption{Example of capsule network predict correctly while without capsule network predict wrongly.}
\label{fig:exp3}
\vspace{-1em}
\end{figure}

\textbf{Case Study on Capsule Network}
We investigate the case that with the capsule network the model predicts correctly, while without the capsule network the model predicts wrongly. As shown in Figure \ref{fig:exp3}, it is hard to answer the question. However, by comparing with the other options, option A will be the best answer. In this case, the evidence is not much helpful for selecting option A. But the capsule network comparing it with other options can help predict ``She wanted her to look at a pretty rock'' as answer.

\end{appendices}
\end{document}